\documentclass[journal]{IEEEtran}

\usepackage{graphicx}
\usepackage{cite}
\usepackage{amsmath}
\usepackage{bm}
\usepackage{amssymb}
\usepackage{tikz}
\usepackage{color}
\usepackage{stfloats}
\usepackage[noend]{algpseudocode}
\usepackage{algorithmicx,algorithm}

\usepackage{booktabs}
\usepackage{hyperref}[colorlinks,linkcolor=black]

\usepackage{subfigure}

\usepackage{amsmath}

\begin{document}
\title{Improving Robotic Grasping Ability \\ Through Deep Shape Generation}
\author{Junnan Jiang$^{1}$, Yuyang Tu$^{2}$, Xiaohui Xiao$^{3}$, Zhongtao Fu$^{4}$, Jianwei Zhang$^{2}$, Fei Chen$^{\dagger 5}$, Miao Li$^{\dagger 1}$

\thanks{This work was supported by the Key Industry Technology Innovation Project of Suzhou (SYG202121) and by the DFG/NSFC SFB/TRR169 (Yuyang Tu, Jianwei Zhang). (Corresponding authors: Fei Chen, Miao Li.)}
\thanks{$^{1}$ Junnan Jiang and Miao Li are with the Institute of Technological Sciences, Wuhan University, Wuhan 430072, China. (e-mail: jiangjunnan@whu.edu.cn, miao.li@whu.edu.cn)}
\thanks{$^{2}$ Yuyang Tu and Jianwei Zhang are with the Department of Informatics, University of Hamburg, Hamburg 20146, Germany}
\thanks{$^{3}$ Xiaohui Xiao is with the School of Power and Mechanical Engineering, Wuhan University, Wuhan 430072, China.}
\thanks{$^{4}$ Zhongtao Fu is with the School of Mechanical and Electrical Engineering, Wuhan Institute of Technology 430205 Wuhan, China}
\thanks{$^{5}$ Fei Chen is with the Department of Mechanical and Automation Engineering, The Chinese University of HongKong, Hongkong 999077, China}
}

\maketitle

\begin{abstract}
 Data-driven approaches have become a dominant paradigm for robotic grasp planning. However, the performance of these approaches is enormously influenced by the quality of the available training data. In this paper, we propose a framework to generate object shapes to improve the grasping dataset quality, thus enhancing the grasping ability of a pre-designed learning-based grasp planning network. In this framework, the object shapes are embedded into a low-dimensional feature space using an AutoEncoder (encoder-decoder) based structure network. The rarity and graspness scores are defined for each object shape using outlier detection and grasp-quality criteria. Subsequently, new object shapes are generated in feature space that leverages the original high rarity and graspness score objects' features, which can be employed to augment the grasping dataset. Finally, the results obtained from the simulation and real-world experiments demonstrate that the grasping ability of the learning-based grasp planning network can be effectively improved with the generated object shapes.
\end{abstract}

\IEEEpeerreviewmaketitle

\section{Introduction}
Grasping is a fundamental ability for robots. Despite significant progress in the area of grasp planning, it is still a difficult task to plan a grasp for unknown objects in general \cite{bohg2013data,li2016dexterous}. During the past decade, data-driven approaches have demonstrated significant advantages over traditional model-based approaches \cite{ferrari1992planning} in the grasp planning area. Since the performance of data-driven approaches is greatly limited by the quality of the available training data, how to improve the quality of training data is one of the crucial questions to improve the grasping ability.

The intuitive idea is to expand the size of the training dataset. Numerous grasping datasets containing more shapes, more annotated grasps and more sensory information have been proposed \cite{jiang2011efficient,calli2015ycb,mahler2017dex,fang2020graspnet,morrison2020egad,gao2022objectfolder} and are trying to include more diverse data for network training. However, except for expensive and time-consuming dataset collection, a well-trained grasp planning network on existing datasets may still fail when facing some special pre-grasped objects. Simply adding a small amount of ``failed objects" to the original large amount of training data is difficult to improve the quality of the training dataset effectively, which further leads to only a small improvement in the grasping ability of the grasp planning network.

\begin{figure}[t]
    \centering
    \includegraphics[width=1\linewidth]{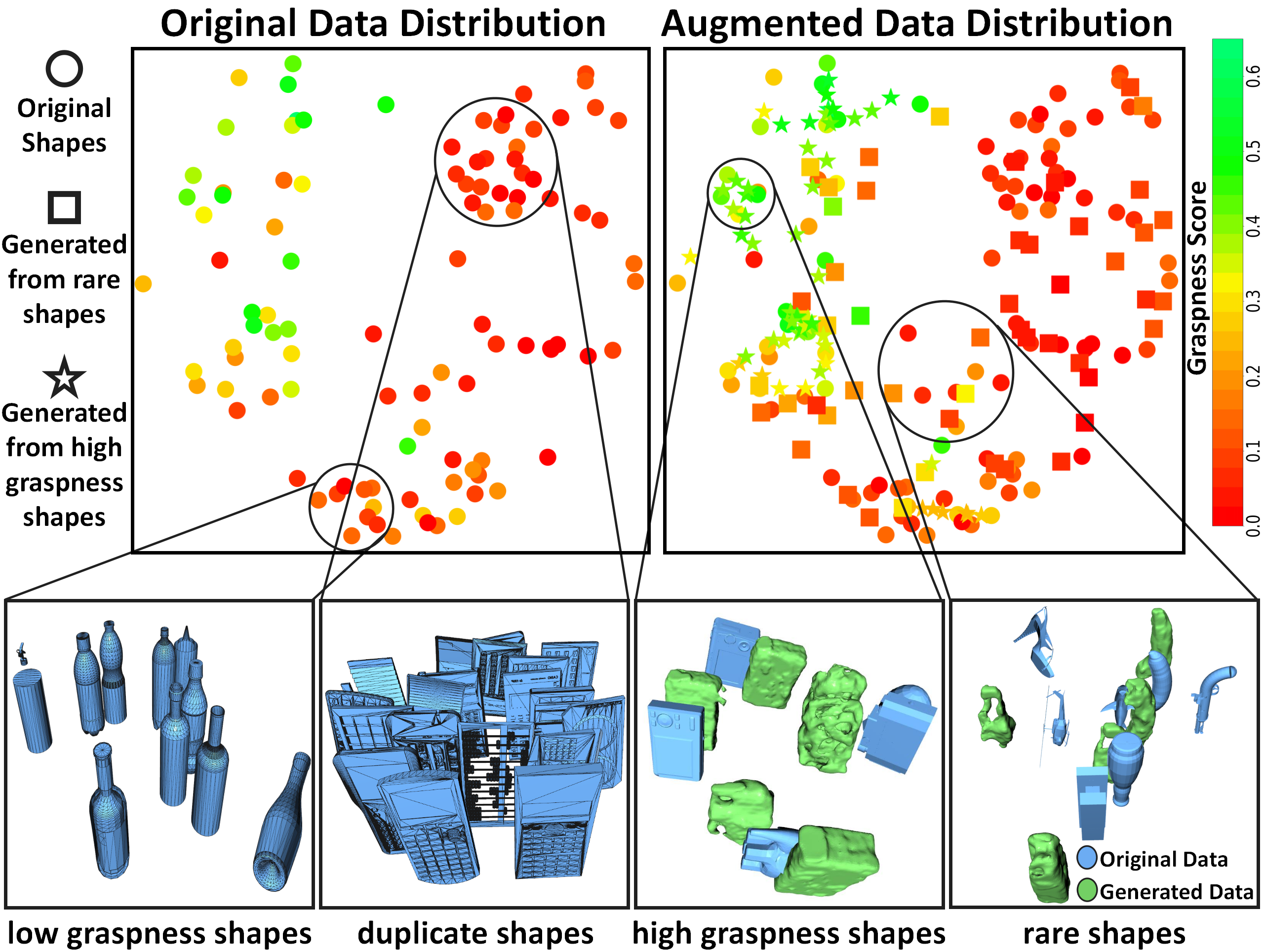}
    \caption{Original and augmented data distribution comparison. We use t-SNE \cite{van2008visualizing} to project the shape feature vectors to a 2D plane, where the Euclidean distances between scattered points represent their features' similarity, and the colors represent their graspness scores. We generate new data that leverages the features of original high rarity and graspness shapes to improve the quality of the dataset.} \label{fig1}
\end{figure}

Another more efficient approach is data augmentation. Similar to computer vision, there are also some data augmentation methods applied in grasping datasets using random image transformation or shape generation \cite{morrison2020learning,tobin2017domain}. However, the effect of these random operations on grasping ability improvement is not certain, and these operations also do not take into account the grasp property. Furthermore, for a grasping dataset shown in Fig.~\ref{fig1}, where the Euclidean distances between scattered points represent their shape similarity and the colors represent their graspness scores defined in \ref{GraspnessScore}, the data distribution may be duplicated in object shapes and uneven in different graspness scores. Randomly augmenting the training data, i.e., generating new data using the features of all data equally, may lead to a larger amount of duplicate shapes and higher uneven distribution of these shapes' graspness scores, which may result in grasp planning network overfitting on these duplicate and uneven data and cannot generalize to other unseen data.

To address these issues, we augment the dataset for the purpose of making it more diverse, and generate data leveraging the features of rare data in the original dataset. In detail, an AutoEncoder-based network is firstly proposed to encapsulate object shape information into a low-dimensional feature space. In this feature space, shapes in the original dataset can be effectively encoded, interpolated and generated. Moreover, in this low-dimensional feature space, two grasp-related metrics are proposed to find the rare data in the grasping dataset. Finally, the features of these rare data are used to generate new shapes that can further improve the quality of the original dataset and thus improve the grasping ability of a pre-defined learning-based grasp planning network. The comparison between the original and augmented data distribution is shown in Fig.~\ref{fig1}: Our newly generated shapes fill the vacant area of the original data distribution and cause a more even graspness score distribution.

The main contributions of this paper are  summarized as:
\begin{itemize}
\item[$\bullet$] An AutoEncoder-Critic network is proposed to map a voxelized shape into a low-dimensional feature space.
\item[$\bullet$] Two grasp-related metrics are proposed to find the rare data in the grasping dataset.
\item[$\bullet$] A systematic approach is proposed to generate new shapes that can improve the grasping ability of a pre-designed learning-based grasp planning network.
\end{itemize}

The remainder of this paper is structured as follows. Section \uppercase\expandafter{\romannumeral2} presents related work in grasping datasets and data augmentation. Section \uppercase\expandafter{\romannumeral3} describes the methodology of our whole data augmentation pipeline, including an object shape encoding method for shape generation, two grasp-related metrics for shape selection, and an augmentation method by the generated object shapes. Section \uppercase\expandafter{\romannumeral4} describes both simulation and real-world experiments, with a final discussion and conclusion in Section \uppercase\expandafter{\romannumeral5}.

\section{Related Work}
\subsection{Grasping Dataset}

With the great success achieved by data-driven grasp planning methods \cite{bohg2013data}, many grasping datasets have been proposed \cite{jiang2011efficient,calli2015ycb,mahler2017dex,fang2020graspnet,morrison2020egad,gao2022objectfolder}. Though existing grasping algorithms can perform well on one dataset, they may still fail when facing unseen objects. Since optimizing the grasping network architectures requires expert experience, it is more desirable to improve the grasping dataset quality and retrain the network, such as expanding the dataset with ``failed objects" or performing some data augmentation tricks. But it is still difficult to answer whether newly added shapes will improve the dataset quality in terms of enhancing the grasping ability. To solve this problem, the EGAD dataset \cite{morrison2020egad} methodically generates shapes with a richer range of shape complexity and grasp difficulty. However, since the shapes generated in EGAD are very different from existing real-world shapes, it is more difficult for a pre-defined network to learn a good grasping policy with an EGAD dataset than with a real-world dataset. This means that the EGAD dataset can only be used for evaluation, but is difficult to apply in real-world scenarios.

\subsection{Data Augmentation}
Data augmentation \cite{cubuk2019autoaugment,shorten2019survey} is a prevalent approach in data-driven approaches because it effectively improves the quality of the dataset and reduces network overfitting problems. Handcrafted methods \cite{imgaug} have been widely used in computer vision, such as shifting, scaling and rotating. Similarly, randomly rotating and cropping images \cite{morrison2020learning} or randomly combining different shapes to generate a new shape \cite{tobin2017domain} can also be used to augment the grasping dataset. However, since it is difficult to evaluate what these random operations bring to the dataset, the effect of these dataset augmentation methods can only be known after retraining the network on the augmented dataset.

With the development of generation methods such as AutoEncoder \cite{bourlard1988auto} and generative adversarial network (GAN) \cite{goodfellow2014generative}, data can be generated more flexibly and even with specific objectives. DeVries et al. \cite{devries2017dataset} augment different-domain data in feature space only with the same AutoEncoder-based network. Wang et al. \cite{wang2019adversarial} generate adversarial grasp objects by evaluating the generated objects' grasp difficulty and regularizing the generation network to generate difficult-to-grasp objects. Mitrano et al. \cite{mitrano2022data} formalize data augmentation as an optimization problem and propose some objective functions to sample better generated data for manipulation tasks. Inspired by these methods, in this work we focus on grasping dataset augmentation and generate data similar to the rare data in a grasping dataset, thereby improving the grasping ability for a pre-defined network.

\section{Object Shape Encoding}
To leverage shapes' features of the original dataset and to ensure that generated shapes are more realistic, we propose an AutoEncoder-Critic (AE-Critic) network. The AutoEncoder (AE) can embed object shapes into a low-dimensional feature space to better generate shapes by interpolation, and the Critic can regularize the generated shapes to be more realistic.

\subsection{Network Architecture}
\begin{figure}[h]
    \centering
    \includegraphics[width=1\linewidth]{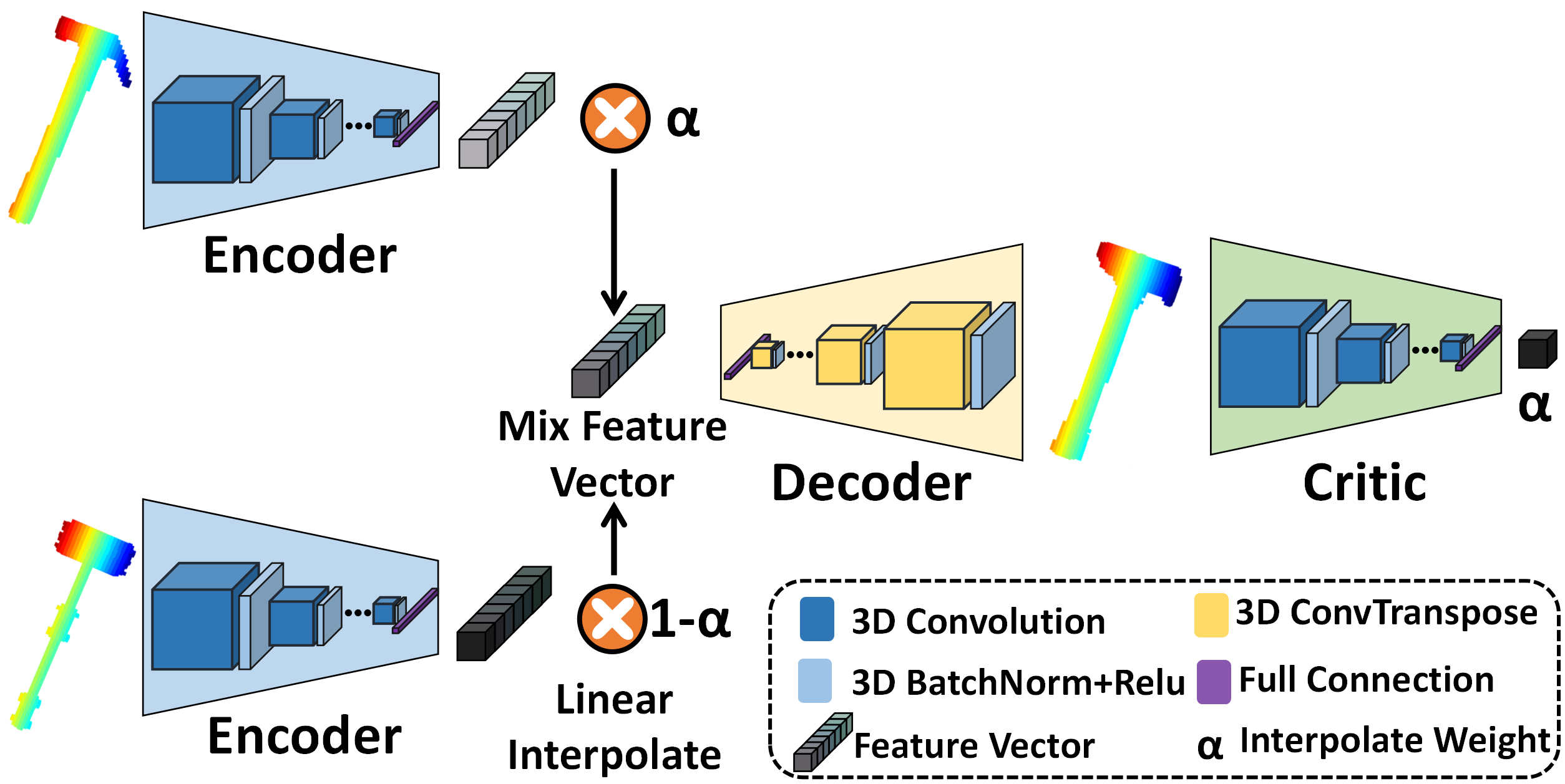}
    \caption{AE-Critic network architecture. The Critic tries to estimate the interpolated weight $\alpha$ corresponding to an interpolated shape and thus can regularize the AutoEncoder (Encoder+Decoder) to generate more realistic interpolated shapes by fooling the Critic to output a smaller interpolated weight.} \label{fig2}
\end{figure}

In order to generate new shapes leveraging the features of the original object shapes, we propose an AE-Critic network and show its structure in Fig.~\ref{fig2}. We use a voxel grid to represent a shape, which maps a shape to a $64\times64\times64$ binary matrix. The AutoEncoder \cite{bourlard1988auto} of AE-Critic contains an Encoder and a Decoder. The Encoder maps a voxel grid $x$ to a 128-dimensional feature vector $z$, containing five 3D convolution layers and two fully connected layers. The convolution layers use $4\times4\times4$ kernel size and two strides, with batch normalization and ReLU layers added in between, mapping the voxel grid to a $512\times4\times4\times4$ sized feature map. The fully connected layers have 32768 and 128 neurons separately, and map the feature map to a 128-dim feature vector. The Decoder mirrors the Encoder, maps a 128-dim feature vector to a $64\times64\times64$ reconstructed voxel grid $\hat{x}$, and contains two fully connected layers and five 3D transposed convolution layers \cite{zeiler2010deconvolutional}. The layer configurations are the same as the Encoder. Using AutoEncoder, we can generate new shapes leveraging the original shapes' features by changing their feature vectors, like interpolating between real samples. In detail, based on the formula $z_{mix}=\alpha z_1+(1-\alpha) z_2$, we can obtain $z_{mix}$ by interpolating two feature vectors $z_1$, $z_2$, which are encoded by two shapes $x_1$, $x_2$, with the interpolated weight $\alpha$. Then we decode the mixed feature vector $z_{mix}$ to generate an interpolated shape $\hat{x}_{\alpha}$. Theoretically, by changing the interpolated pairs $x_1$, $x_2$ and weights $\alpha$, we can generate infinite interpolated shapes $\hat{x}_{\alpha}$.

In addition, to make the shapes generated by interpolation more realistic, we add a Critic inspired by Berthelot et al. \cite{berthelot2018understanding}, which aims to estimate the interpolated weight $\alpha$ corresponding to an interpolated shape $\hat{x}_{\alpha}$, and to regularize the training process of the AutoEncoder. In detail, the AE is trained to generate shapes to fool the Critic to output a smaller interpolated weight, which means that the Critic is more willing to consider the interpolated input as a non-interpolated original shape. Therefore, more realistic shapes that are similar to the original shapes can be generated. The structure of the Critic is similar to the Encoder, only with one more fully connected layer to map the 128-dim feature vector to a 1-dim interpolated weight.

\subsection{Network Training}
The Critic is trained to minimize the loss function given in Eq.~(\ref{eq1}), its first term trains the Critic $C$ to recover interpolation weight $\alpha$ from interpolated voxel grid $\hat{x}_\alpha$ with mean square loss. For the second term, $\gamma$ is a scalar hyperparameter and $\hat{x}$ is a reconstructed voxel grid by AutoEncoder from an original voxel grid $x$. This enforces the Critic $C$ outputs $0$ for non-interpolated inputs, which means the mixture of the original and reconstructed voxel grid in shape space but not in feature space, thus making the Critic's training process more stable.
\begin{equation}
\begin{aligned}
L_C=||C(\hat{x}_\alpha)-\alpha||^2~+||C\left\{ \gamma x+(1-\gamma)\hat{x} \right\}||^2
\end{aligned}
\label{eq1}
\end{equation}

The AutoEncoder is trained to minimize the loss function given in Eq.~(\ref{eq2}), where $L_B$ represents binary cross-entropy loss and $\lambda$ is a scalar hyperparameter to balance the magnitude of the two loss terms. Its first term trains the AutoEncoder to generate a reconstructed voxel grid $\hat{x}$ similar to the original voxel grid $x$, the same as the original AutoEncoder training process. The second term, serving as a regularization term, encourages the interpolated voxel grid $\hat{x}_\alpha$ to fool the Critic to output 0, which means that the Critic considers the generated interpolated voxel grids to be realistic non-interpolated voxel grids. And this can finally regularize the AutoEncoder to generate more realistic voxel grids.

\begin{equation}
L_{(E,D)}=L_B(\hat{x},x)+\lambda ||C(\hat{x}_\alpha)||^2
\label{eq2}
\end{equation}

After the training process, for each object represented as a voxel grid, our AE-Critic network can map it to a 128-dim feature vector. To better visualize the distribution of all the feature vectors, we use t-SNE \cite{van2008visualizing} to project feature vectors to a 2D plane. The distribution of shapes can be seen in Fig.~\ref{fig1}. As shown, it is clear that similar shapes are located close to each other. Moreover, all the shapes are not uniformly distributed in the space and there are many vacant areas, which means no shapes are located there. These areas can be filled by interpolated shapes. The next chapter will explain in detail which vacant areas need to be filled to improve the grasping ability of a given dataset.

\section{Data Augmentation for Robotic Grasping}

The purpose of improving the quality of the grasping dataset to improve the grasping ability is to allow the grasp planning network to learn more diverse data. Therefore, we first define rarity and graspness metrics for shapes, where the higher the metric score, the rarer the data. Then, through the AE-Critic network, new shapes are generated using the features of high-scoring data, which are further used to augment the original dataset. The whole grasping dataset augmentation pipeline is shown in Fig.~\ref{fig3}.

\subsection{Shape Rarity and Graspness Metrics}
\noindent\textbf{Shape Rarity:}

We assume that rare shapes are those whose features are distinct from others. Thus, we use outlier detection \cite{breunig2000lof} to evaluate each feature vector. For one shape which is rarer, the score of outlier detection is higher. In detail, we use the Euclidean distance $dist$ between two feature vectors to measure their similarity. For one shape $O$, we select its k-nearest shapes $N_k(O)$, and the local reachability density $D$ can be defined by Eq.~(\ref{eq3}), which is the reciprocal of the mean of distances between feature vectors from shape $O$ to its $k$ neighbors. This can be used to measure the density around each shape; higher scores indicate greater density.

\begin{equation}
\small
D(O)=\frac{1}{\frac{1}{k}\sum_{P\in N_k(O)}dist(O,P)} 
\label{eq3}
\end{equation}

With the local reachability density metric $D$, we can compute the score of local outlier factor $R$ for each shape's rarity score by Eq.~(\ref{eq4}). It is the average ratio of the $D$ of each shape O to the $D$ of its k nearest neighbors P. A lower object O's $D$ score and higher $D$ scores of its neighbors will result in a higher $R$ score, indicating that shape O is rarer.
\begin{equation}
\small
R(O)=\frac{1}{k}\sum_{P\in N_k(O)}\frac{D(P)}{D(O)} 
\label{eq4}
\end{equation}

\newpage
\noindent\textbf{Shape Graspness:}\label{GraspnessScore}
\begin{figure}[t]
    \centering  \includegraphics[width=1\linewidth]{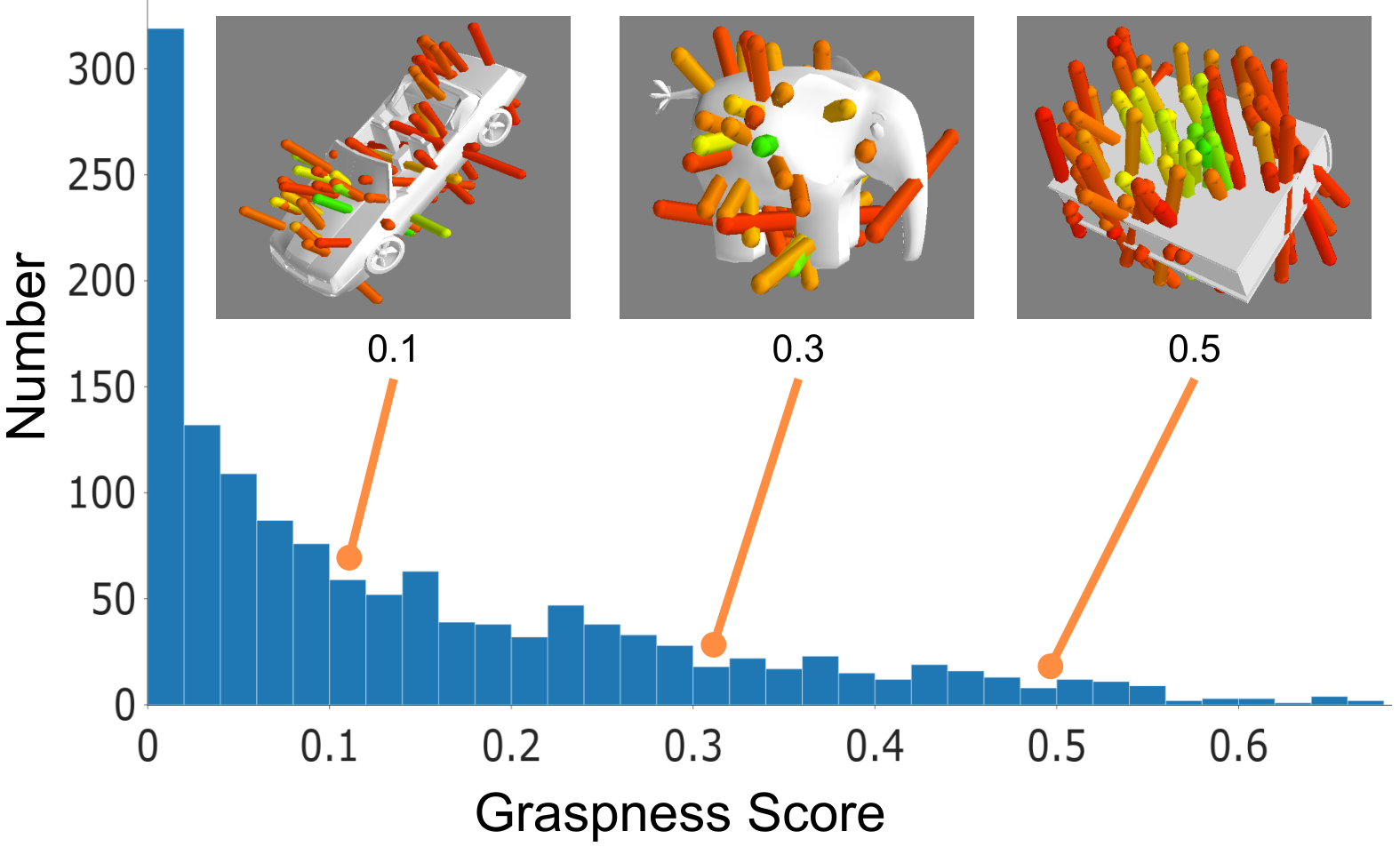}
    \caption{The graspness score histogram of all objects in the 3dnet dataset \cite{wohlkinger20123dnet}. Three objects with different graspness scores are shown above, each cylinder representing an antipodal grasp. The color of the cylinder indicates the grasp quality of each grasp, ranging from red to green.} \label{fig4}
\end{figure}

\begin{figure*}[b]
\centering
\includegraphics[width=\linewidth]{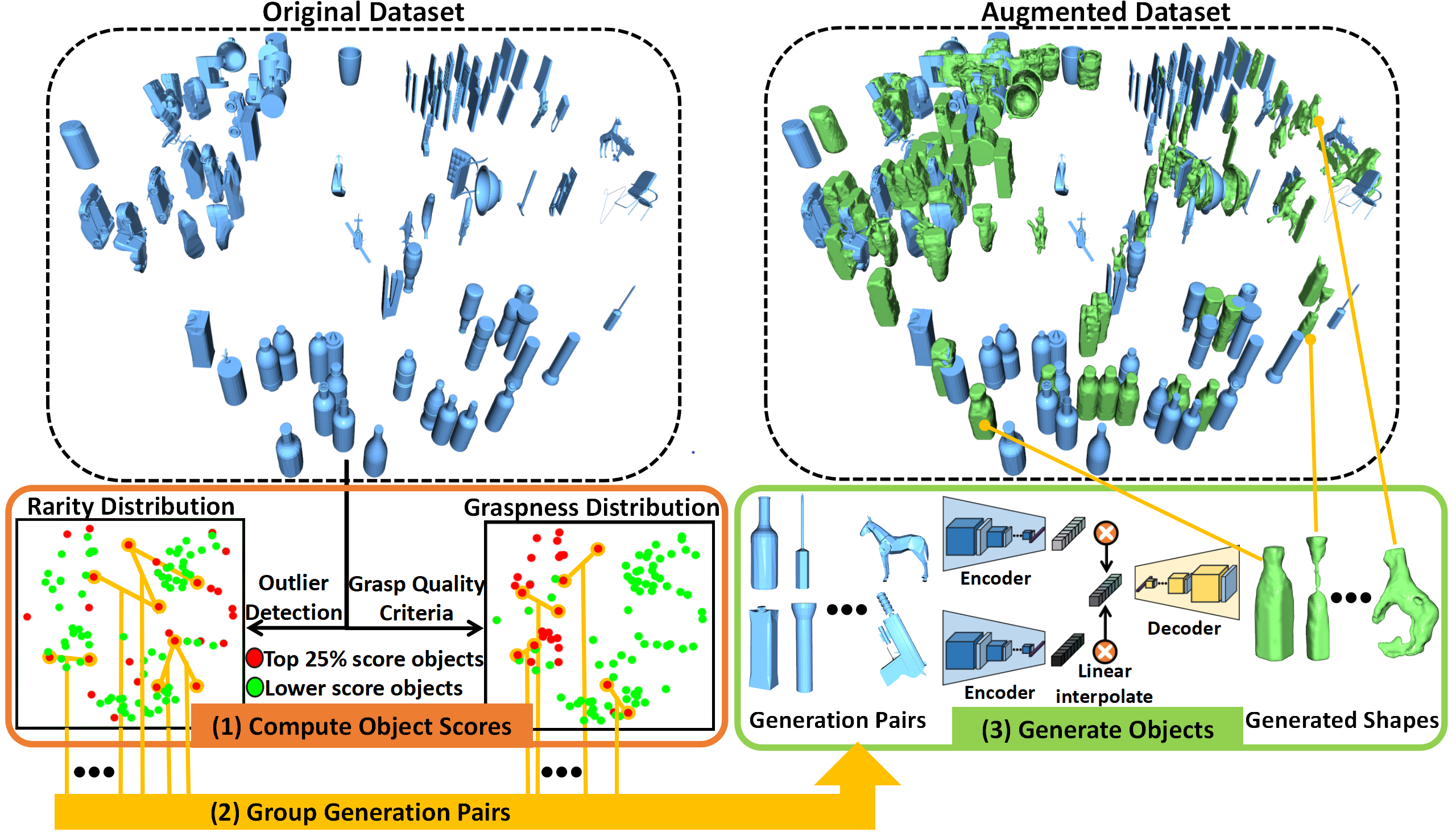}
\caption{The whole pipeline of shape generation for grasping dataset augmentation. The original shapes' rarity and graspness scores are firstly computed through outlier detection and grasp-quality criteria. Then, every two high-scoring nearby objects are grouped as a generation pair. Finally, the AE-Critic network is used for shape generation through interpolation between two shapes' feature vectors.}
\label{fig3}
\end{figure*}

We define a shape’s graspness score as the level of difficulty to find a stable grasp for an object. Firstly, using the Dex-Net analytical grasp planner \cite{mahler2017dex}, a number of antipodal grasps on a shape's surface can be sampled. Then, we use robust Ferrari-Canny \cite{ferrari1992planning} to compute each grasp quality $Q$: 
\begin{equation}
\begin{aligned}
\small
Q&=\mathop{min}\limits_{w} LQ(w)\\
\end{aligned}
\label{eq5}
\end{equation}
where $LQ$ is a local quality metric that measures how efficiently a given wrench $w$ can resist disturbances given applied forces $f$ and the approximated friction cone $FC$:

\begin{equation}
\begin{aligned}
\small
LQ(w)&=\mathop{max}\limits_{f} \frac{||w||}{||f||}\\
&\text{s.t.} ~~f\in FC
\end{aligned}
\label{eq6}
\end{equation}

Finally, we use a threshold of 0.002 \cite{kappler2015leveraging} for the grasp quality to distinguish whether a grasp is successful or not, and define the proportion of successful grasps of an object to all its sampled grasps as its graspness score. The lower the graspness score, the harder the object is to grasp. Fig.~\ref{fig4} is the histogram of graspness scores of all objects in the 3dnet dataset \cite{wohlkinger20123dnet}. The lack of objects with a high graspness score verifies that they are insufficient and need to be generated.

\subsection{Shape Generation}

Based on the AE-Critic network and two defined metrics, we can finally generate shapes leveraging the features of insufficient data to augment the dataset. The overall pipeline is shown in Fig.~\ref{fig3}.

After calculating the rarity and graspness scores of all objects, we only select every two objects whose scores are higher than $t\%$ of all scores in each metric as a generation pair. Then we will have to avoid the feature vectors of shapes in generation pairs being too close, causing shapes to be duplicated, or too far apart, which would cause the intermediate properties to disappear. As this would mean that the intermediate interpolated shape's properties are not similar to the generation pair's properties, we group the nearest $N$-th to $(N+K)$-th neighbors into generation pairs. With these generation pairs, we linearly interpolate with interpolation weight $\alpha$ between each generation pair's feature vectors and decode the mixed feature vector to a newly generated shape. The generated shapes, represented in a voxel grid, are converted to a triangle mesh representation using marching cubes \cite{lewiner2003efficient} and smoothing \cite{vollmer1999improved}.

Up to this point, we leverage the features of shapes with high rarity and graspness score to generate new shapes and get a higher quality grasping dataset. The generated number of augmented shapes depends on the parameters $t$, $N$, $K$ and $\alpha$.

\section{Experiments}

\subsection{Experiment Setup}\label{experiments-setup}

All our experiments are based on the 3dnet dataset \cite{wohlkinger20123dnet} and the GQ-CNN \cite{mahler2017dex} grasp planning algorithm. Since our augmentation method only expands the amount of shapes, other grasping datasets and grasp planning networks can also be used. We randomly select 1000 shapes as the training dataset and 363 shapes as the test dataset. Considering the network is needed to classify the uneven distribution of successful and failed grasps, we use the Average Precision (AP) score on the test dataset to measure the performance of a network. To evaluate the augmentation effect in real-world applications, we also set up a grasping system with a Franka Emika Panda robot arm and an Intel Realsense D435 depth camera. 

\subsection{Critic Regularization Effect}

\begin{figure}[h]
    \small
    \centering
    \includegraphics[width=1\linewidth]{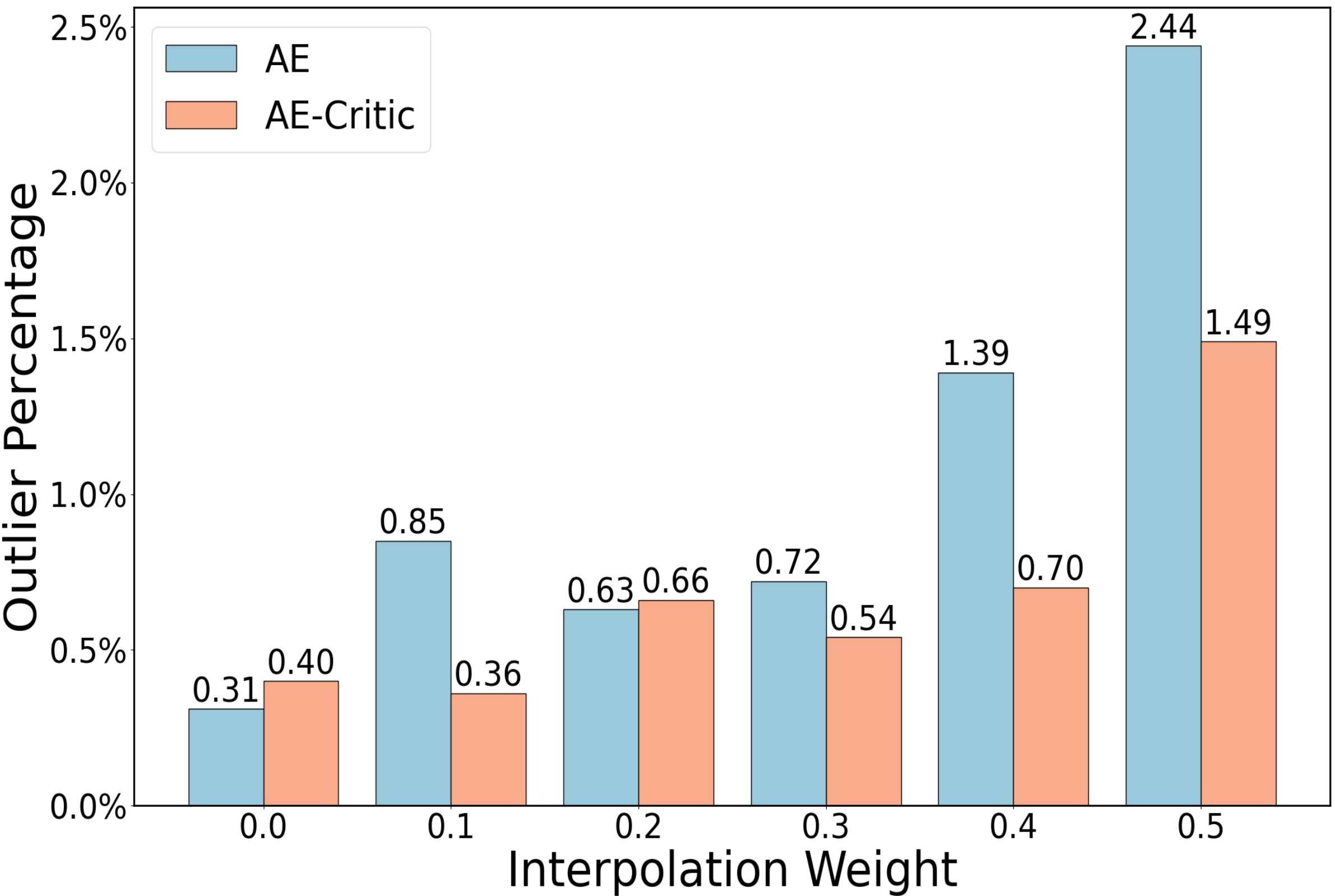}
    \caption{Generated shapes' outlier percentage comparison between AE and AE-Critic network on different interpolated weights. Due to the symmetry of interpolation, we only generate shapes with interpolated weights in the range $[0,0.5]$.} \label{outliercomparison}
\end{figure}

To compare the effect of Critic regularization on shape generation, we evaluate the generated shapes' completeness by AE and AE-Critic networks. In detail, we first train the AE network on 3dnet using the Adam \cite{kingma2014adam} optimizer with a 0.001 learning rate and use the trained AE parameters as a pre-trained network for AE-Critic, and train AutoEncoder and Critic in AE-Critic with a 0.0001 and 0.001 learning rate, respectively. Then,  we perform DBSCAN \cite{ester1996density}, a point cloud clustering method, for each shape to obtain all their clusters. All points in one cluster are contiguous, and the cluster with the highest number of contiguous points is considered to be the major part of a shape, while the points in the other clusters are considered to be outlier points. The percentage of outlier points to all points in a shape is used to evaluate its completeness; the lower the outlier percentage, the more complete the shape is. Finally, we generate 409 shapes with different interpolated weights from 200 randomly selected shapes in the 3dnet dataset, and calculate the percentage of the outlier in the generated shapes. The outlier percentages of two networks based on different interpolated weights are shown in Fig.~\ref{outliercomparison}.

\begin{figure}[h]
    \small
    \centering
    \includegraphics[width=1\linewidth]{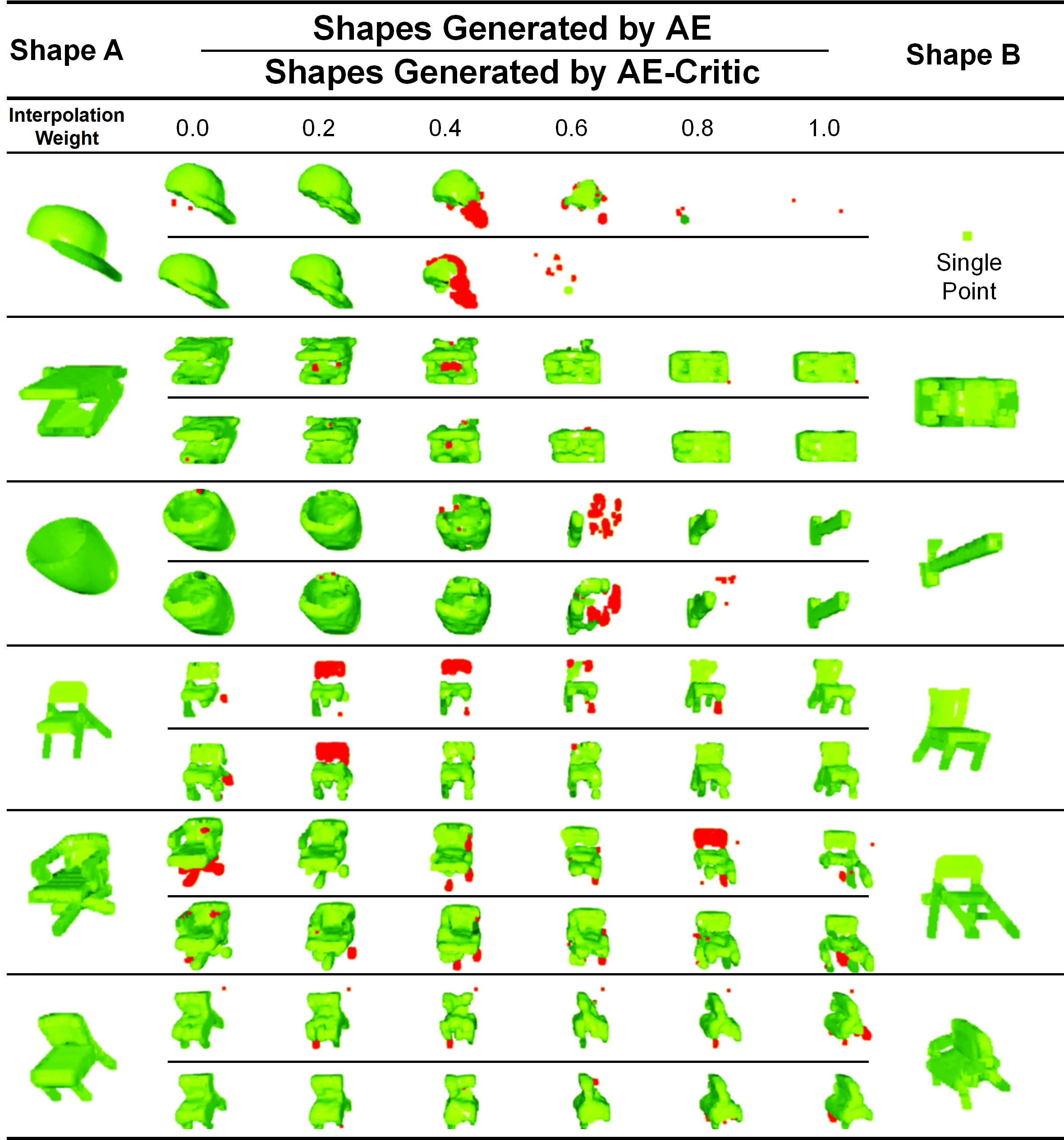}
    \caption{Example of interpolated data from 3D-Net \cite{wohlkinger20123dnet}, produced by AE and the AE-Critic network. The red indicates the outlier points and the green indicates the major part of the shape. In each set of interpolated objects, the top row is generated by AE and the bottom row is generated by AE-Critic.} \label{fig5}
\end{figure}

The results show that higher interpolated weights lead to higher outlier percentages, but the Critic regularization method can reduce the percentage of outlier points in the generated objects. The generated objects are shown in Fig.~\ref{fig5}, with the major part of the shapes in green, and their outlier points in red. And in each set of interpolated shapes, the top row is generated by AE, and the bottom row is generated by AE-Critic.

\subsection{Augmentation Ratio and limitation}
\begin{figure}[h]
    \centering
    \includegraphics[width=1\linewidth]{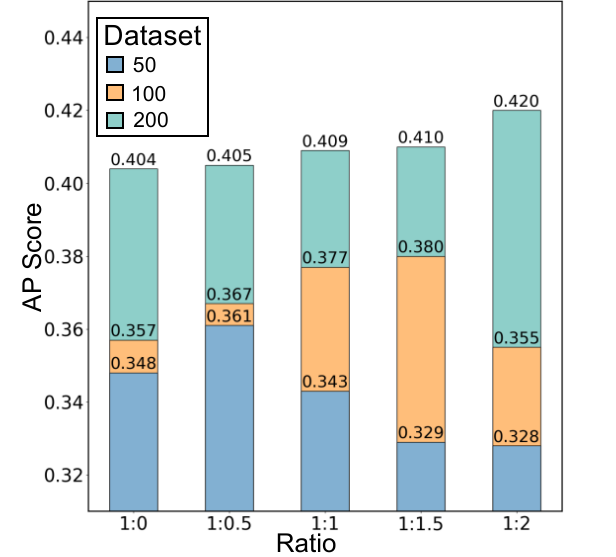}
    \caption{15 GQ-CNN networks trained on 15 augmented datasets and their corresponding AP scores on the same test dataset. The 15 datasets are augmented by different augmentation ratios and the amount of original datasets.} \label{fig7}
\end{figure}

To find the optimal augmentation ratio and augmentation limitation, we randomly select 50, 100, and 200 shapes from 1000 training data, and generate new shapes with 1:0, 1:0.5, 1:1, 1:1.5, and 1:2 augmentation ratios. This means that for 50 original shapes, 0, 25, 50, 75, and 100 generated shapes similar to both in high rarity and graspness score shapes are used for augmentation, the same as 100 and 200 shapes. Then the whole 15 augmented datasets are used for GQ-CNN training, and the 15 trained GQ-CNN networks are tested through the same test dataset mentioned in Section \ref{experiments-setup}, and their corresponding AP scores are shown in Fig.~\ref{fig7}.

Although the results show that our augmentation methods can improve the accuracy of the network on the test dataset, the augmentation ratios allowed by different amounts of data are different. The 50, 100, and 200 data achieve the highest AP values at 1:0.5, 1:1.5, and 1:2, respectively. This means that blindly increasing generated data will lead the network overfitting to the generated data and cause a bad performance on the original dataset.

\subsection{Improvement from Generated Data}\label{augment-experiment}

\begin{figure}[h]
\begin{center}
\includegraphics[width=\linewidth]{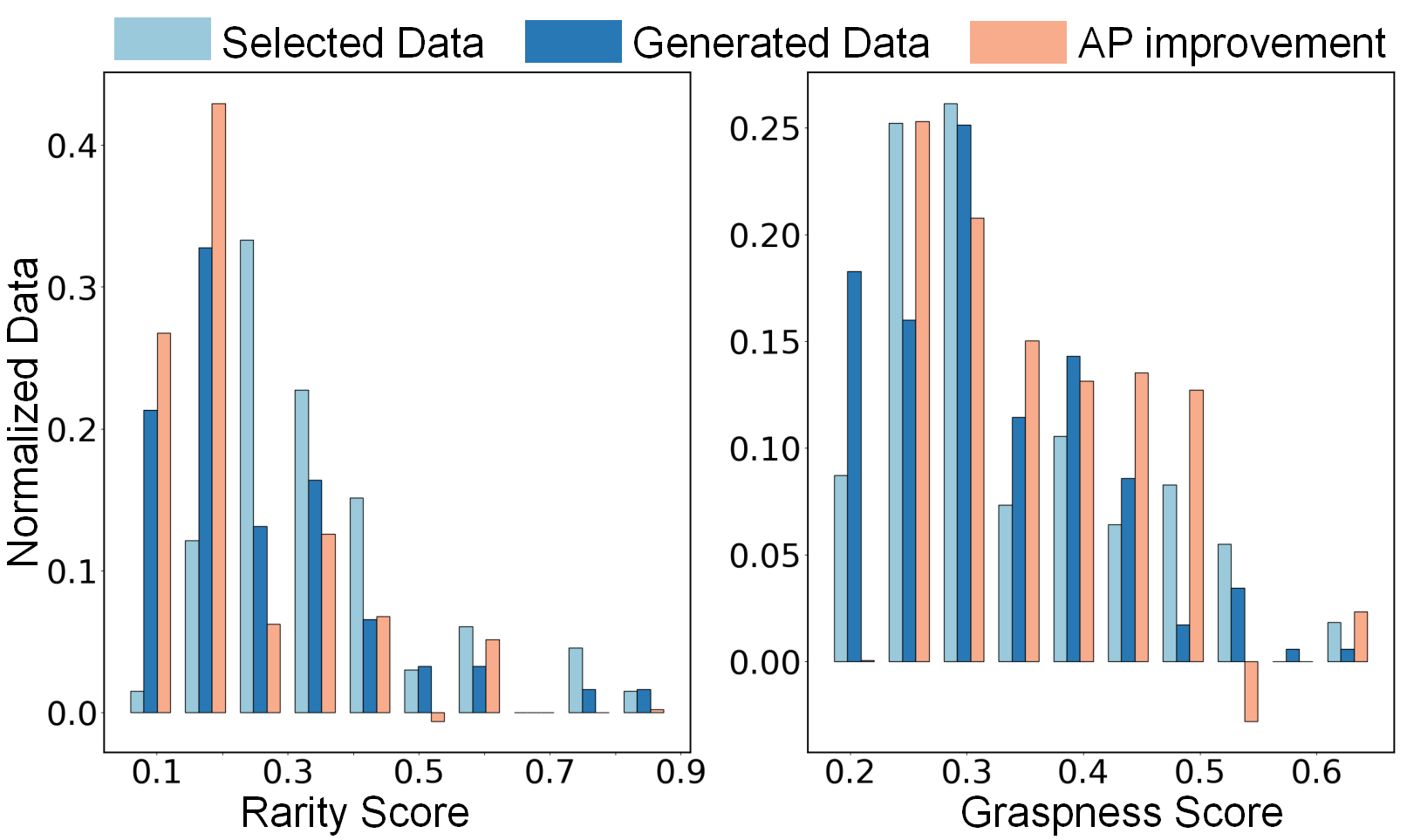}
\end{center}
\caption{The histograms between selected high-scoring data, generated data and AP improvement on the test dataset with different rarity or graspness scores.}
\label{fig6}
\end{figure}

To see the detailed changes brought from our generated data, we compare the correlation between the selected data for data generation, the newly generated data, and the overall network AP improvement on the test dataset caused by training with the augmented data. Specifically, selected data is the data with the top 25\% rarity or graspness scores from the randomly selected 200 shapes in the 3dnet dataset. Generated data is the data generated by leveraging the features of the selected data and using them for data augmentation. 190 and 219 data are generated from the selected high-scoring rarity and from graspness data separately. Both the selected and generated data's distribution histogram of rarity and graspness scores are computed for visualization. We also calculate the AP improvement value of each object with an amount of 363 in the test dataset mentioned in Section A. The AP improvement refers to the improvement of the network on the test dataset after training the GQ-CNN \cite{mahler2017dex} with the augmented data. We sort them into different rarity and graspness score intervals and calculate the average AP score of objects in each score interval. Thus, we can plot the distribution histogram of AP improvement relative to rarity and graspness scores. For the convenience of visualization, all histograms normalize the total number of their distribution to 1 and are plotted together in Fig.~\ref{fig6}.

The histograms show that more selected data will result in more generated data with the same rarity or graspness score, which means that the generated data has the same property as the original selected data to a certain extent. And the generated data at the same time will lead to a greater AP score improvement.

\subsection{Real-world Validation}

\begin{figure}[b]
    \centering
    \includegraphics[width=1\linewidth]{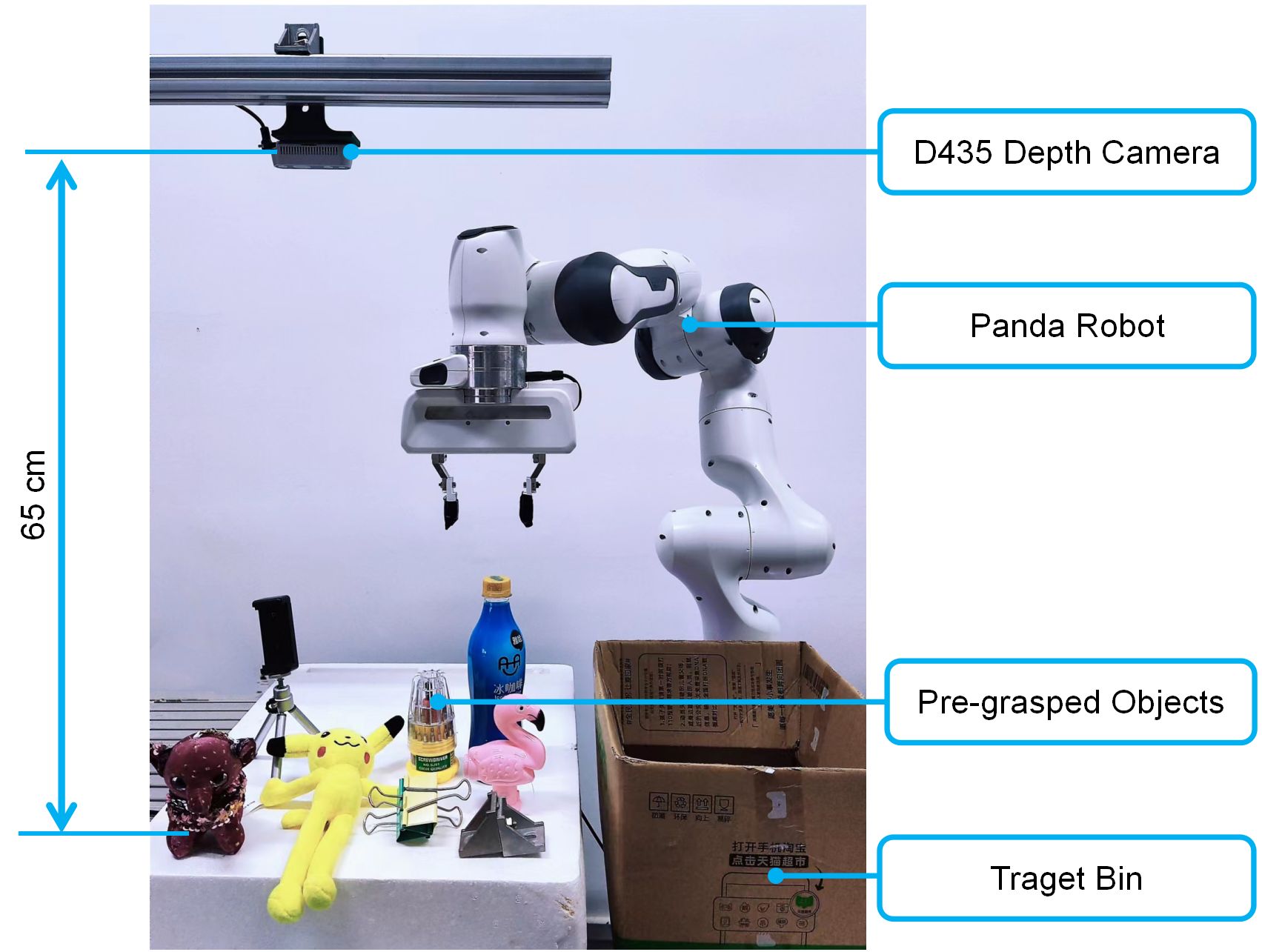}
    \caption{The robotic grasping system.} \label{fig9}
\end{figure}

\begin{figure*}[t]
\centering
    \subfigure[Grasp success rate comparison]{\includegraphics[width=0.48\textwidth]{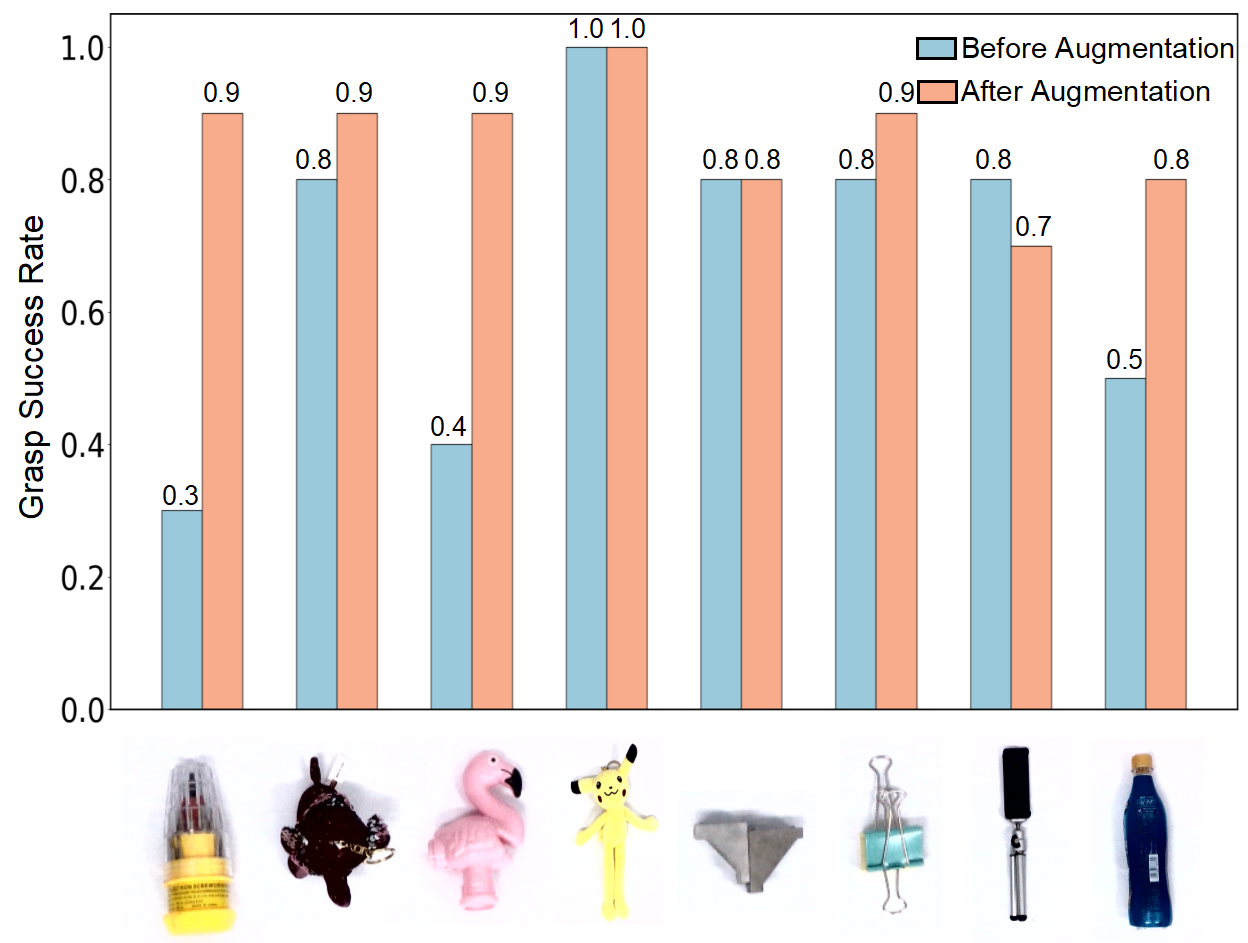}}
    \subfigure[All grasp attempt results]{ \includegraphics[width=0.48\textwidth]{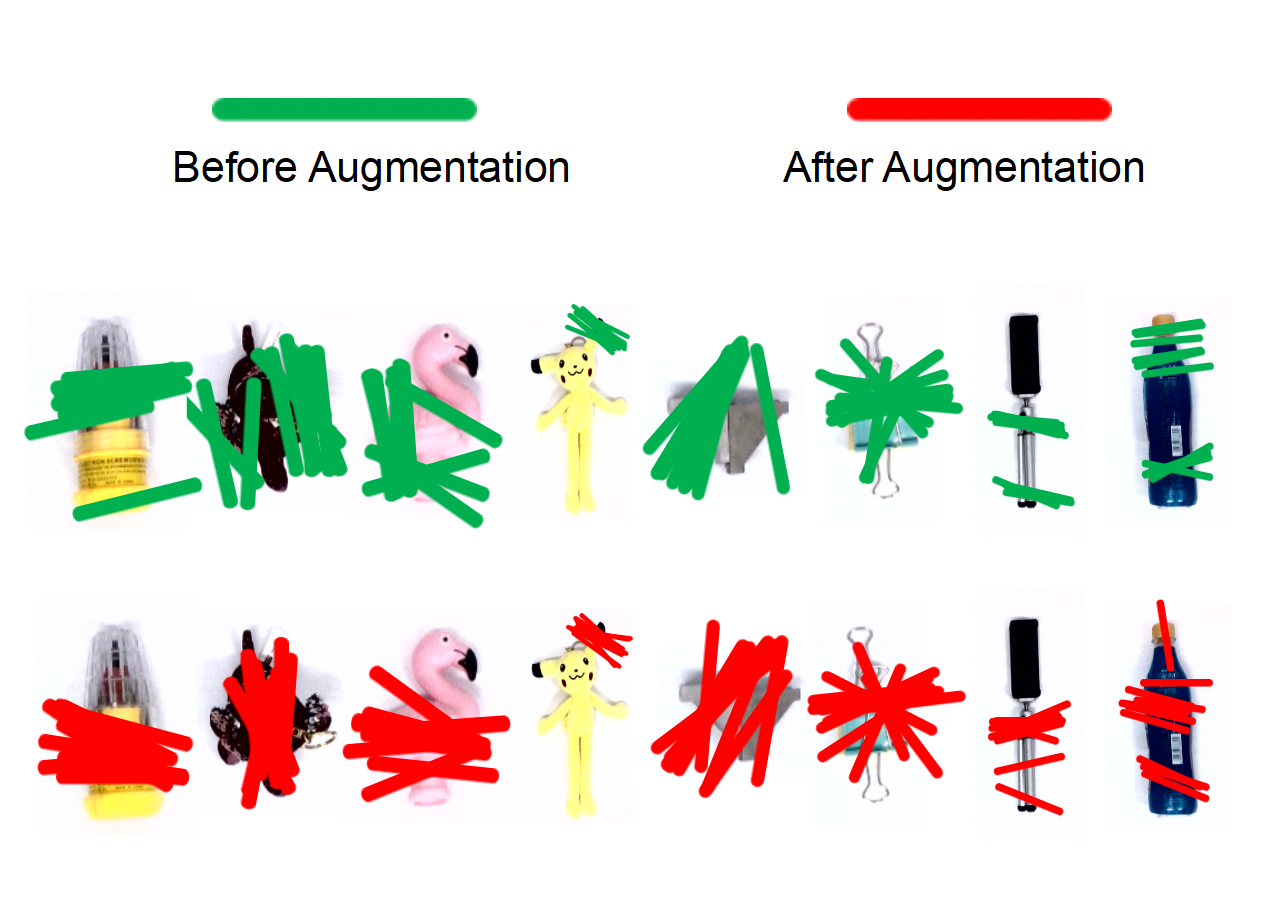}}
\caption{Real-world experiment results. Fig. 10a compares the success rate of each object’s 10 grasp attempts between, before, and after the augmentation GQ-CNN network. \cite{mahler2017dex}. The average grasp success rate increases from 68\% to 86\%. Fig~10b shows each object's 10 grasp attempts' results produced separately before and after the augmentation GQ-CNN network. Green and red lines indicate the grasp attempt results before and after the augmentation GQ-CNN network.}
\label{real-robot-result}
\end{figure*}

\begin{figure}[h]
    \centering
    \includegraphics[width=1\linewidth]{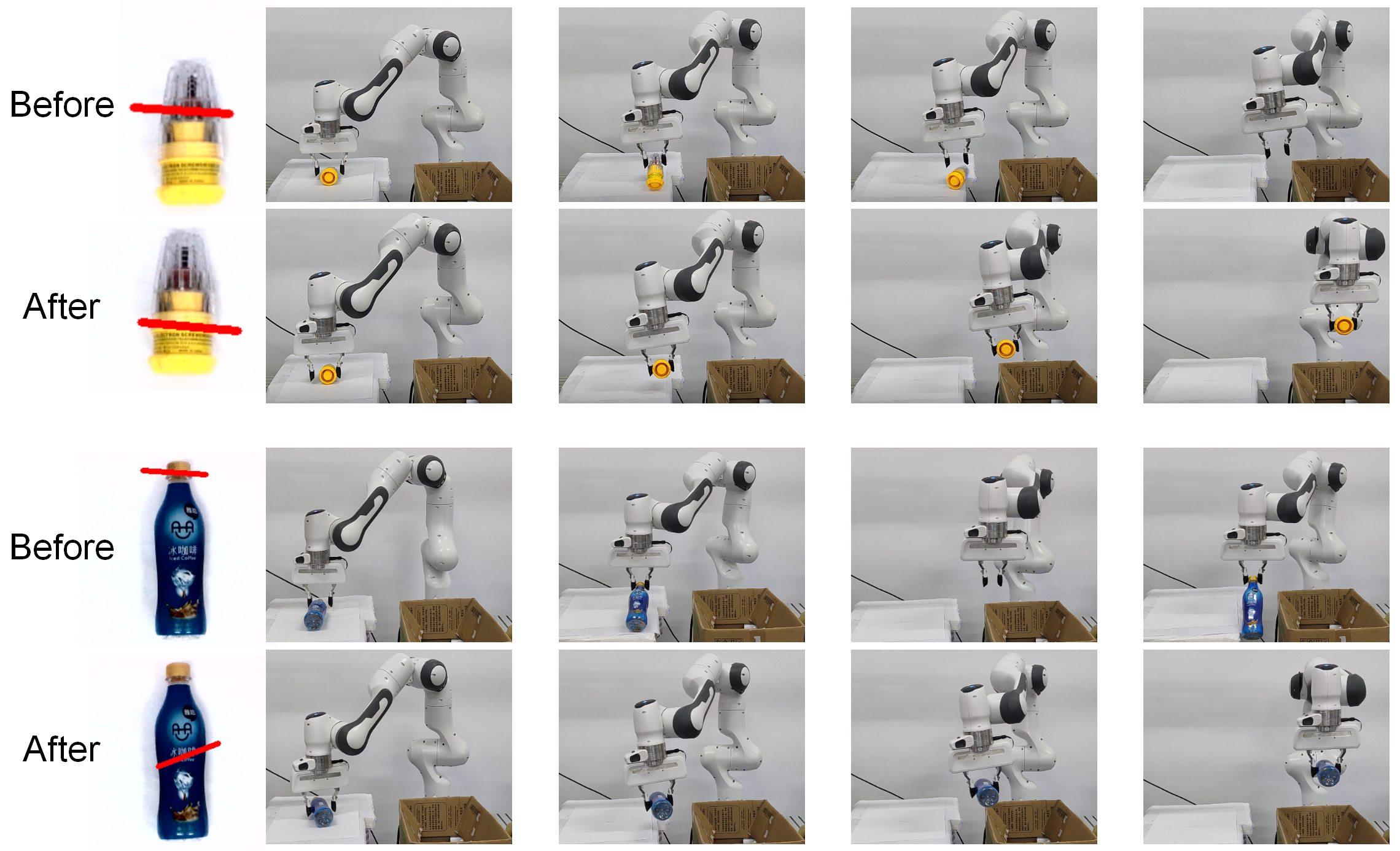}
    \caption{Grasp process comparison between before and after augmentation GQ-CNN network \cite{mahler2017dex}.} \label{GraspProcess}
\end{figure}

To validate the augmentation effect in the real-world applications, we augment 200 original shapes with 409 generated shapes. Both original and generated shapes are the same in Section \ref{augment-experiment}. Then two GQ-CNN \cite{mahler2017dex} networks are trained on the before and after augmentation dataset, and deployed in the grasping system shown in Fig.~\ref{fig9}. The D435 depth camera is fixed 65cm above the grasping platform and eight everyday objects are selected for grasping. A Panda robot performs 20 grasp attempts for each object, ten attempts before and ten after the augmentation GQ-CNN network. For each grasp attempt, a depth image is captured from a fixed viewpoint and the object is placed in the same pose. A grasp is considered successful only when the object is grasped and placed in the target bin. The ten times grasp success rate for each object before and after the augmentation GQ-CNN network is shown in Fig.~\ref{real-robot-result}, and the grasping process is shown in Fig.~\ref{GraspProcess}. All the experimental videos are available at \href{https://youtu.be/Pn6tpSVu5aU}{https://youtu.be/Pn6tpSVu5aU}. Experimental results show that the average grasp success rate increases from 68\% to 86\% using our augmentation method, and validates our augmentation method in real-world scenarios.

\section{Discussion and Conclusion}
\subsection{Discussion}
Although our generated shapes can improve the network's grasping ability in both simulation and real-world experiments, there are still some limitations to this paper. First, in terms of shape generation, either the network structure, shape representation method or feature vector dimension may not be optimal, and a better generation method may result in more realistic and diverse generated shapes, thus further improving the quality of the grasping dataset. Second, in terms of shape selection metrics, the definition and calculation of metrics are not unique. For example, the size and orientation of objects are not taken into account in this paper, and real experimental results can also be used for the calculation of the graspness metric. Finally, shape generation is now only used once for data augmentation, but the shape generation process can be a lifelong process, which means we can continuously use the features of grasp-failed objects for subdivision interpolation and shape generation. By learning more and more shapes, the grasping algorithm may gradually improve its capabilities, and this lifelong learning method is what we hope to investigate more in future work. 

Meanwhile, compare to our initial conference paper \cite{jiang2022learning}, we also investigate the regularization effect of the Critic network and the augmentation method effect on real robots in this paper. And both supplementary experiments have proved the effectiveness of our methods.

\subsection{Conclusion}
In this paper, we present a systematic pipeline for grasping dataset augmentation. Objects are encoded into feature vectors using the AE-Critic network, and generated objects, which are generated by leveraging the features of original high rarity and graspness score objects, are used to augment the original grasping dataset. Experimental results show that our generated data improves the quality of the original grasping dataset, and thus improves the ability of the pre-designed learning-based grasp planning network.

\bibliographystyle{IEEEtran}
\bibliography{reference}

\end{document}